\documentclass[a4paper]{svproc}
\PassOptionsToPackage{hyphens}{url}\usepackage{hyperref}
\usepackage{amsmath}
\usepackage{graphicx}

\usepackage{subcaption}
\usepackage{xcolor}

\usepackage[normalem]{ulem}
\useunder{\uline}{\ul}{}

\begin{document}
\mainmatter              % start of a contribution
\title{Bilevel Optimization for Planning through Contact: A Semidirect Method}
\titlerunning{}  % abbreviated title (for running head)
\author{Benoit Landry\inst{1} \and Joseph Lorenzetti\inst{1}, Zachary Manchester\inst{1} \and Marco Pavone\inst{1}}
\authorrunning{Benoit Landry et al.} % abbreviated author list (for running head)
\tocauthor{Benoit Landry, Joseph Lorenzetti, Zachary Manchester, Marco Pavone}
\institute{Stanford University, Stanford CA 94305, USA,\\
\email{\{blandry,jlorenze,zacmanchester,pavone\}@stanford.edu}}

\maketitle              % typeset the title of the contribution

\begin{abstract}
Many robotics applications, from object manipulation to locomotion, require planning methods that are capable of handling the dynamics of contact. Trajectory optimization has been shown to be a viable approach that can be made to support contact dynamics. However, the current state-of-the art methods remain slow and are often difficult to get to converge. In this work, we leverage recent advances in bilevel optimization to design an algorithm capable of efficiently generating trajectories that involve making and breaking contact. We demonstrate our method's efficiency by outperforming an alternative state-of-the-art method on two benchmark problems. We moreover demonstrate the method's ability to design a simple periodic gait for a quadruped with 15 degrees of freedom and four contact points.
\end{abstract}
\section{Introduction}
Trajectory optimization is a well-established method to generate motion plans for dynamical systems. One important application for such methods is motion planning and control for systems with contact dynamics, which are ubiquitous in real-world robotic systems. However, trajectory optimization for systems undergoing contact has proven to be inherently difficult due to the drastically changing dynamical behavior arising from making and breaking contact. For this reason, methods have been proposed which generally fall under several categories. Those using continuous models, such as spring-damper models, have been proposed to eliminate the discontinuities that make the problem so challenging. However in these methods approximation error is induced and numerical problems can arise due to stiffness in the resulting differential equations. Additionally, approaches using hybrid models \cite{Aceituno-CabezasMastalliEtAl2017} have been introduced which can richly capture the true discontinuous dynamics. However it is well known that trajectory optimization formulations using such models, primarily mixed-integer programs, suffer from computational limitations resulting from a combinatorially large number of mode switching possibilities.

An alternative class of approaches, which is the focus of this work, includes the contact forces as decision variables in a trajectory optimization framework. These methods, referred to as planning ``through contact'', utilize techniques made popular for simulation of contact-driven dynamical systems. The primary advantage of this category of methods is that rich contact behavior can be captured without the combinatorial explosions in complexity seen in hybrid model based methods. However, even though the formulation of such optimization problems is generally straightforward and the resulting problem non-combinatorial, they remain numerically challenging for optimization algorithms to handle. Therefore, the exploration of different formulations of the embedded contact dynamics problem and exploration of different solution techniques are both tasks of high importance.

In this work, we propose a variation of the ``through contact'' method first put forward in \cite{PosaCantuEtAl2014}. Our approach leverages recent advances in bilevel and differentiable optimization in which a lower level optimization problem is embedded within an upper level optimization. A key aspect of our algorithm is that it gets rid of the complementarity constraints usually needed to model \textit{friction} forces (which tend to be the cause of many numerical difficulties) and only requires complementarity constraints relating to non-penetration constraints (i.e. for the normal forces). Instead, we compute friction forces as an embedded optimization problem. Note that variations on the idea of modeling contact through bilevel optimization have been explored in the past \cite{CariusRanftlEtAl2018}. However, our formulation posses several appealing characteristics such as being able to leverage state-of-the-art off-the-shelf solvers like SNOPT and OSQP in order to outperform alternative approaches. Moreover our method avoids linearizing the non-penetration constraint which can easily lead to infeasible trajectories. Finally, unlike approaches relying on iLQR, our direct transcription based method can easily and effectively be parallelized, which is of particular relevance when modeling contact forces as solutions to embedded optimization problems.

% An important note the reader: in this work we use the term direct, indirect and semidirect to refer to the way the \emph{lower} (contact force) problem is solved. Unless specified, the upper problem is always solved using a direct method such as direct transcription. Because the literature does not always acknowledges the bilevel nature of contact dynamics, this might lead to confusion as we might refer to a technique as being indirect (e.g. \cite{PosaCantuEtAl2014}) while the authors, not focusing on the embedded optimization problem use the term direct to refer to their method.

An important note the reader: in this work when discussing approaches to trajectory optimization through contact we use the term ``direct'', ``indirect'', and ``semidirect'' to refer to the way \textit{the contact force is resolved} within the trajectory optimization problem, and not the way the problem itself is solved \textit{overall} (for which we do in fact use a direct method). Specifically, by a ``direct'' method we mean a method which defines the contact forces in the trajectory optimization problem completely as the result of an embedded optimization problem (i.e. via the maximum dissipation principle), and by an ``indirect'' method we mean a method which completely resolves the contact forces through the use of complementarity constraints (which can be interpreted as KKT conditions) \cite{PosaCantuEtAl2014}. We refer to our proposed method as ``semidirect'' because it handles \textit{only the friction forces} using an embedded optimization problem, and the normal forces through complementarity constraints.

\subsection{Statement of Contributions}
The contributions of this paper are as follows. First, by combining different modeling approaches to contact forces, we formulate a ``semidirect'' trajectory optimization method for planning through contact. Second, we demonstrate how our formulation, a specific bilevel optimization problem, is well suited to leverage recent advances in differentiable optimization as well as state-of-the-art quadratic programming and nonlinear programming solvers. Next, we demonstrate how our method, based on direct transcription, can easily and effectively leverage parallelization to offset the cost incurred by its bilevel structure. Finally we provide evidence that the method offers a promising direction of research by demonstrating it is capable of outperforming a well accepted alternative method on a benchmark problem, and also generate contact rich interactions between a complex robot and its environment.

\section{Previous Work}

Here we survey methods related to ``through contact'' trajectory optimization and how they differ from our proposed algorithm.

One of the first instance of such methods \cite{PosaCantuEtAl2014} leverages work on rigid body simulation with contact \cite{StewartTrinkle2000} in order to formulate a direct transcription problem capable of generating trajectories that make and break contact. We differ from this approach mainly in that instead of using complemetarity constraint to enforce contact dynamics, we use the maximum dissipation principle to model the friction forces as an embedded quadratic program, and leverage advances in bilevel optimization to handle the newly formed problem. Note that modeling contact forces as optimization problems using the maximum dissipation principle is not an entirely new idea and has been explored in the past. Notably \cite{DrumwrightShell2010} uses this approach to simulate forward contact dynamics and provides a survey of difference instances of this modeling approach.

Perhaps closest to our method, \cite{CariusRanftlEtAl2018} demonstrated a method for planning through contact that also leverages a contact dynamics formulation derived from the principle of maximum dissipation. However, our method differentiates itself from this work on a few points. First, we use a direct collocation method instead of iLQR (a shooting method), which allows us to parallelize the computationally more demanding aspects of the method. Second, instead of solving the lower problem with a projected gradient descent algorithm, we use an off-the-shelf quadratic program solver OSQP which provides faster convergence through the Alternating Direction Method of Multipliers (ADMM). Third, we recover the gradients of the embedded problem analytically, instead of autodifferentiating the solver itself, which again is more computationaly efficient. Lastly, we preserve the normal force as decision variables in the upper-level problem.

Finally, there has also been some work on using the analytical derivatives of quadratic programs in the context of contact. Notably \cite{AvilaBelbute-PeresSmithEtAl2018} embeds a non-convex quadratic program corresponding to the Linear Complementarity Problem (LCP) described in \cite{StewartTrinkle2000} to learn dynamics involving contact. Our method differs from this work in that we perform nonlinear optimization with an off-the-shelf constrained nonlinear programming solver to solve the upper problem instead of stochastic gradient descent. Moreover, the formulations of our lower problem differ as we do not use the non-convex LCP contact formulation, but rather the convex maximum dissipation optimization problem (from which the LCP formulation can in fact be derived from as KKT conditions).

\section{Contact Dynamics as Optimization Problems}\label{sec:contact-opt}

A powerful way of understanding the nature of contact forces is to reason about them using the principle of maximum dissipation. As we show in this section, by performing a first-order discretization of the system dynamics using a backward Euler integration scheme, it is possible to then model the contact forces as the solution of a quadratic program.

\subsection{Maximum Dissipation}
We start with the rigid body dynamics written in manipulator form
\begin{equation}
    \begin{aligned}
        M(q) \dot{v} + c(q,v) + \lambda = \tau,
    \end{aligned}
\end{equation}
where $q$, $v$ are the configuration and joint velocities of the system. The matrix $M(q)$ is the positive-definite inertia matrix, the function $c(q,v)$ is the dynamics bias (i.e. the Coriolis and potential terms of the manipulator equation combined), $\lambda(q,v)$ denotes the contact forces and $\tau$ represents the control input (usually joint torques for robotic systems). This dynamics equation is then discretized in time using a backward Euler integration scheme to yield
\begin{equation}
    \begin{aligned}
        M(q_{i+1})(v_{i+1} - v_i) + h c(q_{i+1},v_{i+1}) - h \tau_{i+1} + h \lambda_{i+1} = 0, \\
        q_{i+1} = q_i + hv_{i+1}, \\ 
    \end{aligned}
\end{equation}
where $h$ is the discretization time step. 

Next, we define the contact force acting at each contact point as a vector in a frame that is centered at the contact location. The normal force component of the contact force vector at a given point is denoted by $c_n$, and the (tangential) friction forces by $f_x$ and $f_y$. Thus the vector $x = [c_n, f_x, f_y]^T$ defines the contact force in the associated contact point frame. With this definition, the dynamics can be written as
\begin{equation}\label{eq:dyn}
    \begin{aligned}
        M(q_{i+1})(v_{i+1} - v_i) + h c(q_{i+1},v_{i+1}) - h \tau_{i+1} + h J(q_{i+1})^T x_{i+1} = 0. \\
        q_{i+1} = q_i + hv_{i+1}, \\ 
    \end{aligned}
\end{equation}
where $J(q)$ is the Jacobian that maps the contact forces from the contact frame to the joint space. 

In order for us to be able to use the principle of maximum dissipation when determining the contact forces, we now derive an expression for the change in kinetic energy from one time step to the next
% \begin{equation}
%     \begin{aligned}
%         dT &= T_+ - T_-\\
%         &= \frac{1}{2}v_+^T M v_+ - \frac{1}{2}v_-^T M v_-\\
%         &= \frac{1}{2} (v_- - h M^{-1}(J^Tx + c - \tau))^T M (v_- - h M^{-1}(J^Tx + c - \tau)) - \frac{1}{2}v_-^T M v_- \\
%         &= \frac{1}{2}h^2(J^Tx + c - \tau)^T M^{-1} (J^Tx + c - \tau) - h v_-^T(J^T x + c - \tau)
%     \end{aligned}
%     \label{eq:maxdiss}
% \end{equation}

\begin{equation}
    \begin{aligned}
        dT &= T_{i+1} - T_i,\\
        &= \frac{1}{2}v_{i+1}^T M_{i+1} v_{i+1} - \frac{1}{2}v_i^T M_i v_i,\\
        &= \frac{1}{2} (v_i - h M_{i+1}^{-1}(J_{i+1}^Tx_{i+1} + c_{i+1} - \tau_{i+1}))^T M_{i+1} \\
        &\hspace{20mm}(v_i - h M_{i+1}^{-1}(J_{i+1}^Tx_{i+1} + c_{i+1} - \tau_{i+1})) - \frac{1}{2}v_i^T M_i v_i, \\
        &= \frac{1}{2}h^2(J_{i+1}^Tx_{i+1} + c_{i+1} - \tau_{i+1})^T M_{i+1}^{-1} (J_{i+1}^Tx_{i+1} + c_{i+1} - \tau_{i+1}) \\
        &\hspace{20mm} - h v_i^T(J_{i+1}^T x_{i+1} + c_{i+1} - \tau_{i+1}) + \frac{1}{2}v_i^T(M_{i+1} - M_i)v_i.
    \end{aligned}
    \label{eq:maxdiss}
\end{equation}
where $dT$ is the change in kinetic energy, and for notational simplicity we use $M_{i} = M(q_{i})$, $c_{i}=c(q_{i},v_{i})$, and $J_{i}=J(q_{i})$. Note that $dT$ can be expressed quadratically in terms of the contact force $x$. This now allows us to formulate a quadratic program to find the contact force $x$ that maximizes the energy dissipation.

\subsection{Friction Forces as Quadratic Program}
We can treat $q_i$, $v_i$, $\tau_{i+1}$, $M(q_i)$, $M(q_{i+1})$, $c(q_{i+1},v_{i+1})$ and $J(q_{i+1})$ as known quantities, because they are \emph{for the lower solver} in the context of bilevel optimization. Therefore, to find the contact force that provides maximal energy dissipation we seek to minimize the quantity $dT$ in \eqref{eq:maxdiss}. This is equivalent to minimizing the following quadratic function.
\begin{equation}
\frac{1}{2} x_{i+1}^T Q^d_{i+1} x_{i+1} + (r^{d}_{i+1})^T x_{i+1},
\end{equation}
where
\begin{equation}
    \begin{aligned}
    Q^d_{i+1}&= h J_{i+1} M_{i+1}^{-1} J_{i+1}^T, \\
    r^d_{i+1}&=\big(h(c_{i+1} - \tau_{i+1})^T M_{i+1}^{-1}  -  v_i^T\big)J_{i+1}^T.
    \end{aligned}
\end{equation}

Additionally we want to ensure that the contact forces satisfy constraints imposed by a Coulomb friction model. To accomplish this, we first linearize the friction cone, as described in \cite{StewartTrinkle2000}, such that $D \beta = [0,f_x,f_y]^T$ where $D$ is a basis that spans the friction cone and $\beta$ is a vector of non-negative coefficients. We also define $z = [c_n, \beta^T]^T$, and therefore can express the contact force vector $x$ as
\begin{equation}
    \begin{aligned}
        x = Fz, \quad F:=\begin{bmatrix} \hat{n} | D \end{bmatrix},
    \end{aligned}
\end{equation}
where $\hat{n}$ is the unit length normal vector at the contact point. Then, we impose upon the energy dissipation maximization problem the Coulomb friction model constraint $\mu c_n - e^T \beta \geq 0$, where $\mu$ is the friction coefficient and $e = [1,1,\dots,1]^T$.

The resulting computation of the friction force is therefore given by the following quadratic program
\begin{equation}
\begin{aligned}
\beta = \;
& \underset{\beta}{\text{argmin}}
& & \frac{1}{2} z^T F^T Q^d F z + (r^d)^T F z \\
&
& & \mu c_n - e^T \beta \geq 0,\\
&
& & \beta \geq 0,
\end{aligned}
\label{eq:frictionprob}
\end{equation}
where the time index notation has been dropped for clarity. Note that the formulation of our ``semidirect'' method allows us to make the assumption here that the normal force $c_n$ is known and is therefore not a decision variable in \eqref{eq:frictionprob}. This will be discussed in further detail next in Section \ref{sec:trajoptcontact}.

% \begin{equation}
% \begin{aligned}
% \Gamma(q_-,v_-,q_+,v_+,c_n) = \;
% & \underset{\beta}{\text{minimize}}
% & & \frac{1}{2} z^T F^T Q_d F z + r_d^T F z \\
% &
% & & \mu c_n - \mathbf{e}^T \beta \geq 0,\\
% &
% & & \beta \geq 0.
% \end{aligned}
% \label{eq:frictionprob}
% \end{equation}

\section{Trajectory Optimization Through Contact}\label{sec:trajoptcontact}
We now show how to formulate the trajectory optimization problem as a bilevel optimization problem using the results from Section \ref{sec:contact-opt}. As mentioned before, this formulation is ``semidirect'' in the sense that part of the overall contact force (normal forces) are handled via complementarity constraints, and the other part (friction forces) are handled via the maximum dissipation optimization problem given by \eqref{eq:frictionprob}.

Thus in this formulation the decision variables of the trajectory optimization problem are the control inputs $\tau$, configuration variables $q$, joint velocities $v$, and the normal force component $c_n$, but the friction force vector $\beta$ will \textit{not} be a decision variable, as it will be implicitly encoded by the embedded optimization problem.

Note that handling the non-penetration constraints, and hence the normal force, in a lower problem is also possible. However since it is a constraint that is a function of position that would be imposed on decision variables relating to forces, keeping the embedded constraint linear would introduce compounding approximations (double integration). This is not the case for the problem of friction since this later set of constraints are functions of velocity (not position) imposed on decision variables relating to forces.

\subsection{Dynamics Constraints}
For a planning horizon $i = 1,\dots, m$, from the backward Euler discretized manipulator equation \eqref{eq:dyn} with the contact forces in contact space represented by $x=Fz$ where $z=[c_n,\beta]^T$, we have the dynamics constraints
\begin{equation} \label{eq:maneq}
\begin{split}
M_{i+1}(v_{i+1} - v_i) + h c_{i+1} + h J_{i+1}^T Fz_{i+1} = h \tau_{i+1}, \\
q_{i+1} = q_i + hv_{i+1}. \\ 
\end{split}
\end{equation}
The normal force $c_n$ is then constrained by the set of complementarity conditions
\begin{equation}\label{eq:normcomp}
\begin{split}
\phi(q_{i+1}) \geq 0, \\
c_{n,i} \geq 0, \\
c_{n,i} \phi(q_{i+1}) = 0,
\end{split}
\end{equation}
where $\phi(q)$ is a distance function such that $\phi(q)\geq0$ implies non-penetration of the rigid body with its environment.
Finally, the friction force vector $\beta$ is given by \eqref{eq:frictionprob}.

\subsection{Trajectory Optimization}
Given an integrating stage cost function $J(q,v,\tau)$, and problem-specific constraints on the configuration and control defined by $g(q,v,\tau) \leq 0$ and $h(q,v,\tau) = 0$ (i.e. desired initial or final configurations, or actuation limits), the trajectory optimization problem is defined as
\begin{equation}\label{eq:trajopt}
\begin{aligned}
& \underset{q_i,v_i,\tau_i,c_{n,i};i=1\;\ldots m}{\text{minimize}}
& & \sum_{i=1}^{m}{J(q_i,v_i,\tau_i)} \\
& \text{subject to}
& & M_{k+1}(v_{k+1} - v_k) + h c_{k+1} + h J_{k+1}^T Fz_{k+1} = h \tau_{k+1}, \\
&
& & q_{k+1} = q_k + hv_{k+1}, \\
&
& & z_{k+1} = [c_{n,k+1}, \beta_{k+1}]^T, \\
&
& & \phi(q_i) \geq 0, \\
&
& & c_{n_i} \geq 0, \\
&
& & c_{n_i} \phi(q_{i+1}) = 0, \\
&
& & \beta_{k+1} = \underset{\beta_{k+1}}{\text{argmin}} \frac{1}{2} z_{k+1}^T F^T Q^d_{k+1} F z_{k+1} + (r^d_{k+1})^T F z_{k+1} \\
&
& & \hspace{20mm} \mu c_{n,k+1} - e^T \beta_{k+1} \geq 0,\\
&
& & \hspace{20mm} \beta_{k+1} \geq 0. \\
&
& & g(q_i,v_i,\tau_i) \leq 0, \\
&
& & h(q_i,v_i,\tau_i) = 0, \\
\end{aligned}
\end{equation}
where $k = 1,\dots,m-1$.

The above problem forms a nonlinear bilevel trajectory optimization problem that can be solved by off-the-shelf nonlinear solvers as long as special care is taken to handle the friction force embedded problem, as described next in Section \ref{sec:lowerprob}.

\section{Solving the Friction Force Lower Problem}\label{sec:lowerprob}

Now that we have defined the lower problem of our bilevel optimization, namely the quadratic program \eqref{eq:frictionprob} whose solution corresponds to the friction forces at each contact point, we discuss our solution method for it.

There exists many solutions methods capable of handling convex quadratic programs similar to the one described in \eqref{eq:frictionprob}. However, in the context of bilevel optimization problems, additional requirements are placed on the solvers for the embedded mathematical program. First, solutions are needed very quickly and with little overhead since the embedded solver is called frequently by the primary solver working on the upper problem (i.e. for every evaluation of the constraints of the upper problem by the primary solver). In order to address this requirement, we leverage a state-of-the-art quadratic program solver OSQP \cite{StellatoBanjacEtAl2017}, that implements the popular Alternating Direction Method of Multipliers (ADMM) algorithm \cite{BoydParikhEtAl2011}.

Second and perhaps most importantly, gradients of the solution with respect to the parameters of the problem need to be available.

\subsection{Gradients}
In order to provide gradients of the solution, we leverage work in sensitivity analysis \cite{LevyRockafellar1995,FiaccoIshizuka1990,RalphDempe1995}, which has regained traction more recently \cite{AmosKolter2017,AgrawalBarrattEtAl2019,Barratt2018}. The approach relies on using first order optimality conditions and the implicit function theorem in order to define a system that can be solved to recover the gradient of the solution with respect to the problem's parameters. The details of this derivation are outside the scope of this paper and we refer the reader to \cite{AgrawalBarrattEtAl2019} for a thorough treatment of the problem. But for the reader's benefit we include the system that must be solved along with the quadratic program in order to recover the gradients (taken from \cite{Barratt2018}). Given the following quadratic program:

\begin{equation}
\begin{aligned}
& \underset{x}{\text{minimize}}
& & \frac{1}{2}x^T Q(\theta) x + q(\theta)^T x \\
& \text{subject to}
& & G(\theta) x \preceq h(\theta), \\
&
& & A(\theta) x = b(\theta),
\end{aligned}
\end{equation}
where $\theta$ is the parameter vector for which we are interested in getting the gradients with respect to. We can then compute the desired gradient $D_{\theta}x^*$ by solving
\begin{equation}
      \Pi
  \begin{bmatrix}
    D_{\theta}x^* \\
    D_{\theta}\lambda^*\\
    D_{\theta}\nu^*
 \end{bmatrix}
= z,
\end{equation}
where
\begin{equation*}
 \Pi =
 \begin{bmatrix}
    Q & G^T & A^T \\
    \textbf{diag}(\lambda^*)G & \textbf{diag}(Gx^* - h) & 0 \\
    A & 0 & 0
 \end{bmatrix}, \quad  z =
 \begin{bmatrix}
    dQx^* + D_{\theta}q + dG^T \lambda^* + dA^T \nu^* \\
    \textbf{diag}(\lambda)(dGx^* - D_{\theta}h) \\
    dAx^* - D_{\theta}b
 \end{bmatrix}, \\
\end{equation*}
and where $x^*$, $\lambda^*$, $\nu^*$ are the optimal solution for the primal, and dual variables ($\lambda$ is the dual for the inequality constraint and $\nu$ is the dual for the equality constraints). Note that the system must often be solved using a least-squares method.

\section{Parallelization}
One of the main drawback of our approach is that even though the size of the problem (in number of variables and constraints) for the upper problem is reduced, constraint evaluation is now significantly more expensive. Indeed, instead of evaluating a series of inequalities like we would normally be required to, constraint evaluation with our approach requires us to solve a mathematical program. However, because we perform our trajectory optimization using a direct transcription method (and not say iLQR as in \cite{CariusRanftlEtAl2018}), we can trivially run our constraint evaluation in parallel. In our implementation, each dynamic constraint (one per successive knot point) can be evaluated in parallel as its own thread. The performance gain is therefore dependent on the number of knot points used. When using a computer with sufficiently many cores, this parallelization can help offset the cost incurred by the more computationally expensive constraint evaluation as shown in section \ref{subsec:parallelres}.

\section{Results}
We now present some results to empirically demonstrate and validate our proposed approach. First, we demonstrate our algorithm on a smaller problem and compare its efficiency with an indirect method which is considered state-of-the-art \cite{PosaCantuEtAl2014}. For thoroughness, we also benchmark a slightly larger problem consisting of a hopping robot. Next, we use our algorithm to design a `stepping forward' motion for a quadruped that can then be used in repetition as a gait.

\subsection{Implementation Details}
We implement our approach using the Julia programming language \cite{BezansonKarpinskiEtAl2012}. Specifically, the rigid body dynamics (excluding contact dynamics) are computed using the package RigidBodyDynamics.jl \cite{Koolencontributors2016} and we implement the analytical gradient of the embedded quadratic program inside the ForwardDiff.jl framework \cite{RevelsLubinEtAl} (allowing for easy integration with other constraints). Additionally, for the upper problem we use the popular sequential quadratic programming (SQP) solver SNOPT \cite{GillMurrayEtAl2005}, which is often regarded as the most performant solver for trajectory optimization. For the embedded problem, we use the solver OSQP (as described in Section \ref{sec:lowerprob}). Our experiments were run on a 16-core 3.0 GHz CPU with 32Gb of memory. All of of our code is made available at \url{https://github.com/blandry/Bilevel.jl}.

\subsection{Performance Benchmarks} \label{subsec:perfbench}

First we solve a simple planning problem involving sliding a single rigid body on a surface to a target location, where the rigid body is modeled as having a single point of contact with the surface. The initial position of the object is chosen such that it is not in contact with the surface, so that the solver must also determine the contact time. Additionally, a final position constraint is defined, the final velocity is constrained to be zero, and there is no control input along the trajectory. The problem therefore corresponds to finding the right initial velocity to throw the object such that it slides across a surface to the target position. Note that this task is more challenging than it might appear at first glance, since it requires the trajectory to include transitions from non-contact to sliding friction, and then to sticking friction.

Here we compare the result of solving this trajectory optimization problem using our proposed semidirect method against the indirect method presented in \cite{PosaCantuEtAl2014}. The solver SNOPT was also used for the approach in \cite{PosaCantuEtAl2014}, and in both methods was allowed to run until optimality (with a tolerance of $10^{-5}$). Note that even for this simple example, the indirect method required us to introduce slack variables on the complementarity constraints related to friction (a common trick to handle these constraints). Notably however, since our semidirect method handles the friction contact constraints via the embedded problem given by \eqref{eq:frictionprob} no slack variables were necessary.

\begin{figure}
\centering
% \begin{subfigure}{\textwidth}
%   \centering
%   \includegraphics[width=.75\linewidth]{slidingbox_x.eps}
% \end{subfigure}
% \begin{subfigure}{1\textwidth}
%   \centering
%   \includegraphics[width=.75\linewidth]{slidingbox_z.eps}
% \end{subfigure}
\begin{subfigure}{.49\textwidth}
  \centering
  \includegraphics[width=\textwidth]{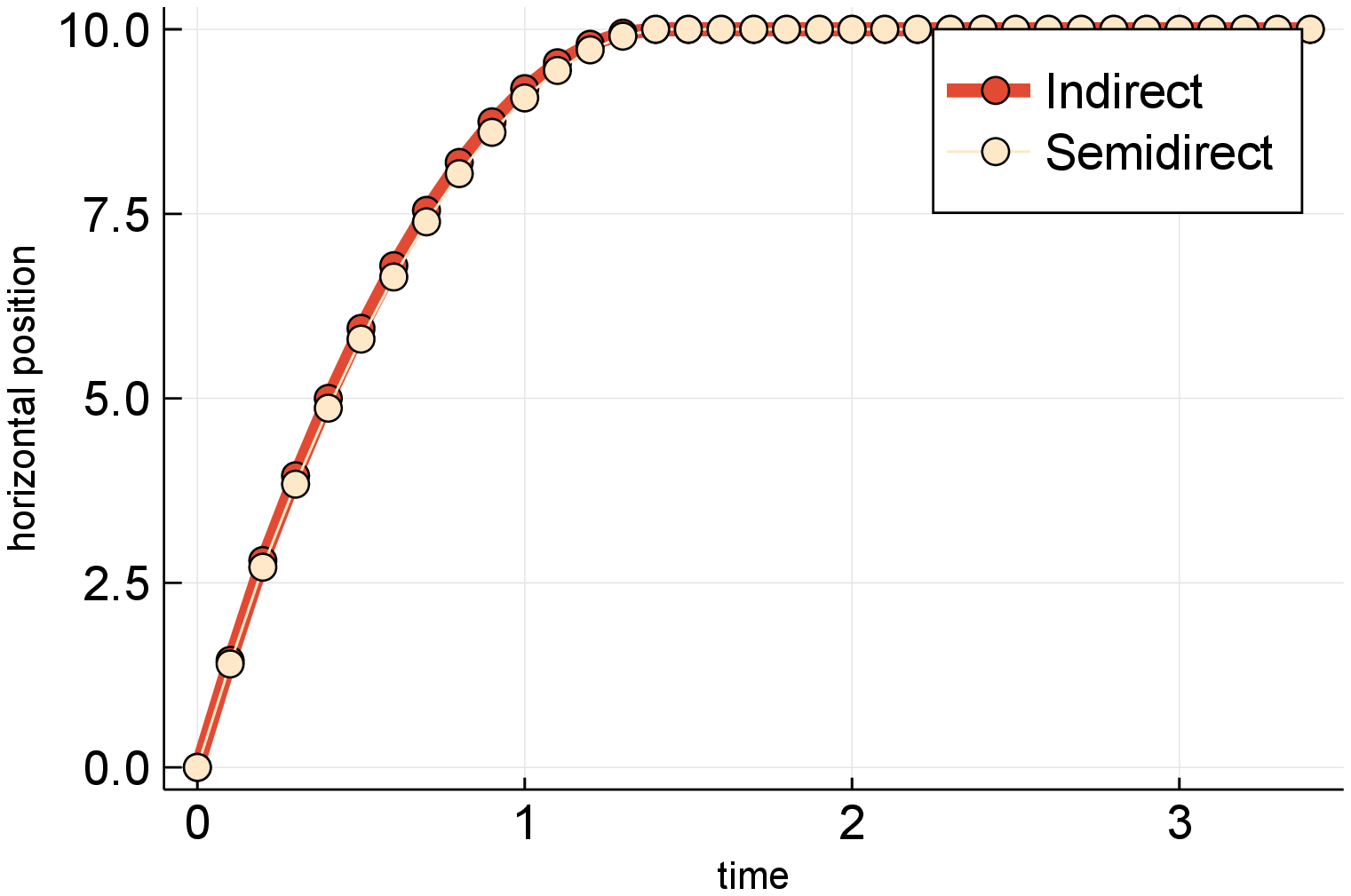}
\end{subfigure}
\begin{subfigure}{.49\textwidth}
  \centering
  \includegraphics[width=\textwidth]{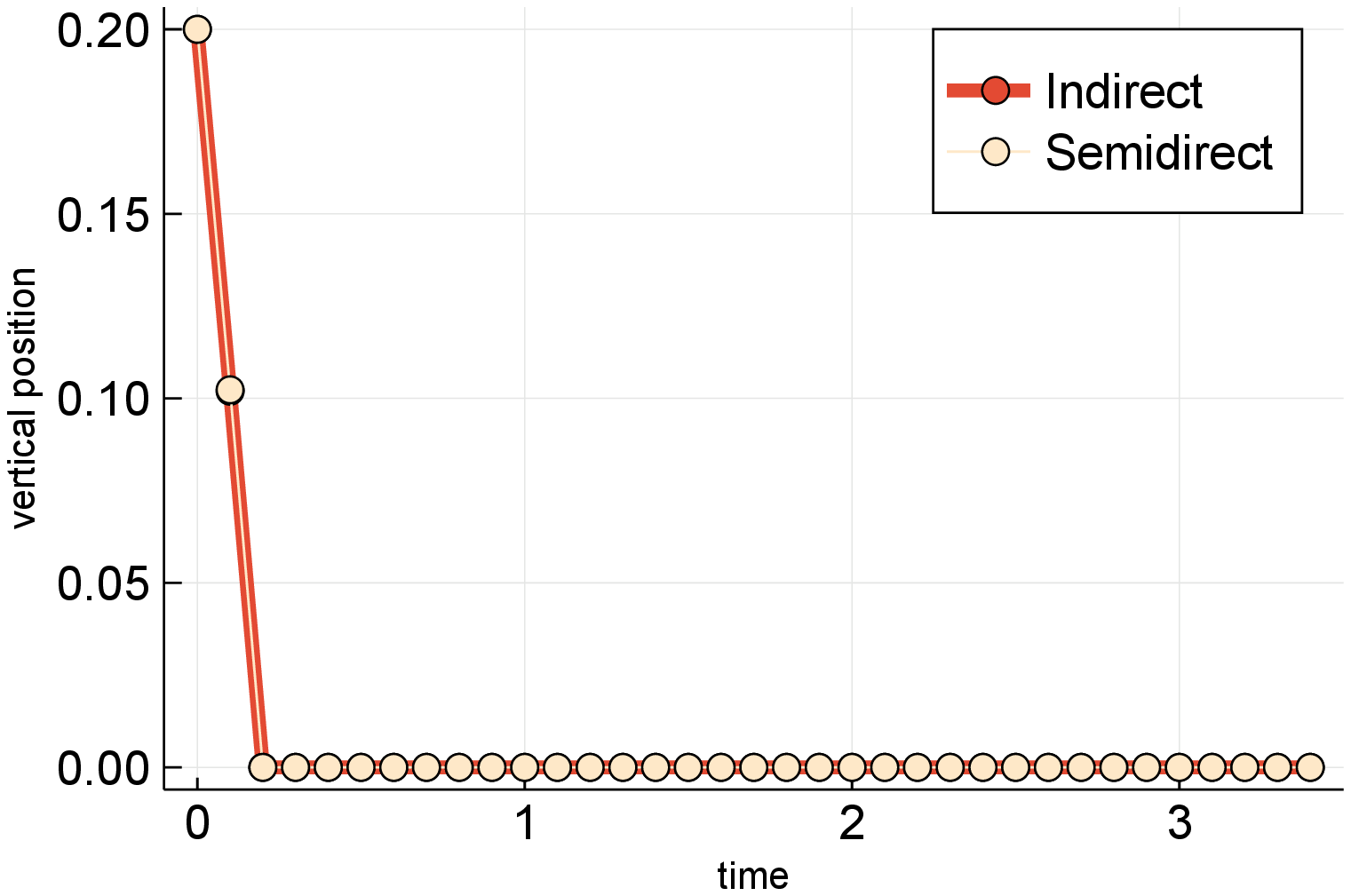}
\end{subfigure}
\caption{Comparison of the horizontal (left) and vertical (right) position trajectories of the rigid body's center of mass resulting from our proposed semidirect method and the indirect method presented in \cite{PosaCantuEtAl2014}, when applied to the problem described in Section \ref{subsec:perfbench}. Units are in seconds and meters.}
\label{fig:slidingbox}
\end{figure}

From the trajectory shown in Figure \ref{fig:slidingbox} and the results presented in Table \ref{tab:slideingbox}, we can see that our method not only recovers a solution of comparable quality, but it does so in less time. For example, for the largest problem we report here, our semidirect method benchmarked (over several averaged samples) at 1.258 seconds, while the indirect method required more computation time, at 2.247 seconds. Note also that for this example, the semidirect method only required SNOPT to solve a problem with 380 variables while the indirect one contained 652.
\begin{table}[]
\centering
\begin{tabular}{|c|c|c|c|c|}
\hline
\textbf{\# Knot points} & \multicolumn{2}{c|}{\textbf{\# Variables (upper problem)}} & \multicolumn{2}{c|}{\textbf{Mean solve time (s)}} \\ \cline{2-5} 
 & \textbf{Indirect} & \textbf{Semidirect - ours} & \textbf{Indirect} & \textbf{Semidirect - ours} \\ \hline
10 & 177 & 105 & 0.070 & {\ul 0.054} \\ \hline
15 & 272 & 160 & 0.232 & {\ul 0.140} \\ \hline
20 & 367 & 215 & 0.456 & {\ul 0.229} \\ \hline
25 & 462 & 270 & 0.686 & {\ul 0.359} \\ \hline
30 & 557 & 325 & 1.036 & {\ul 0.524} \\ \hline
35 & 652 & 380 & 2.247 & {\ul 1.258} \\ \hline
\end{tabular}
\vspace{0.25cm}
\caption{Comparison of optimization problem size (for the upper problem) and solve time (in seconds) between our proposed semidirect method and the indirect method presented in \cite{PosaCantuEtAl2014}, when applied to the problem described in Section \ref{subsec:perfbench}.}
\label{tab:slideingbox}
\end{table}

\begin{figure*}[ht]
    \setlength{\fboxsep}{0pt}%
    \setlength{\fboxrule}{1pt}%
    \centering
    \begin{subfigure}[t]{.19\textwidth}
        \frame{\includegraphics[trim={10cm 17cm 10cm 9cm},clip,width=\textwidth]{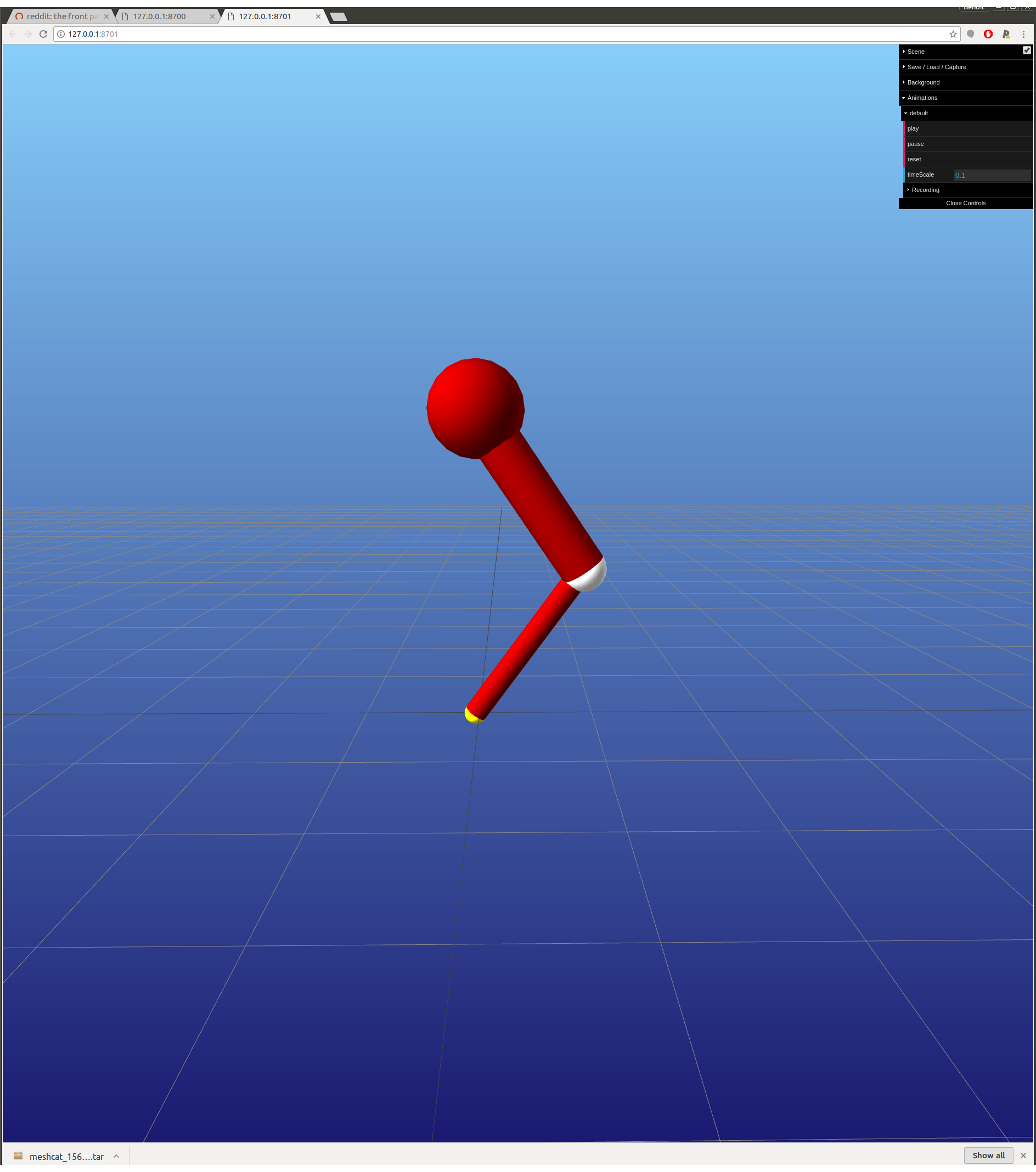}}
    \end{subfigure}
    \begin{subfigure}[t]{.19\textwidth}
        \frame{\includegraphics[trim={10cm 17cm 10cm 9cm},clip,width=\textwidth]{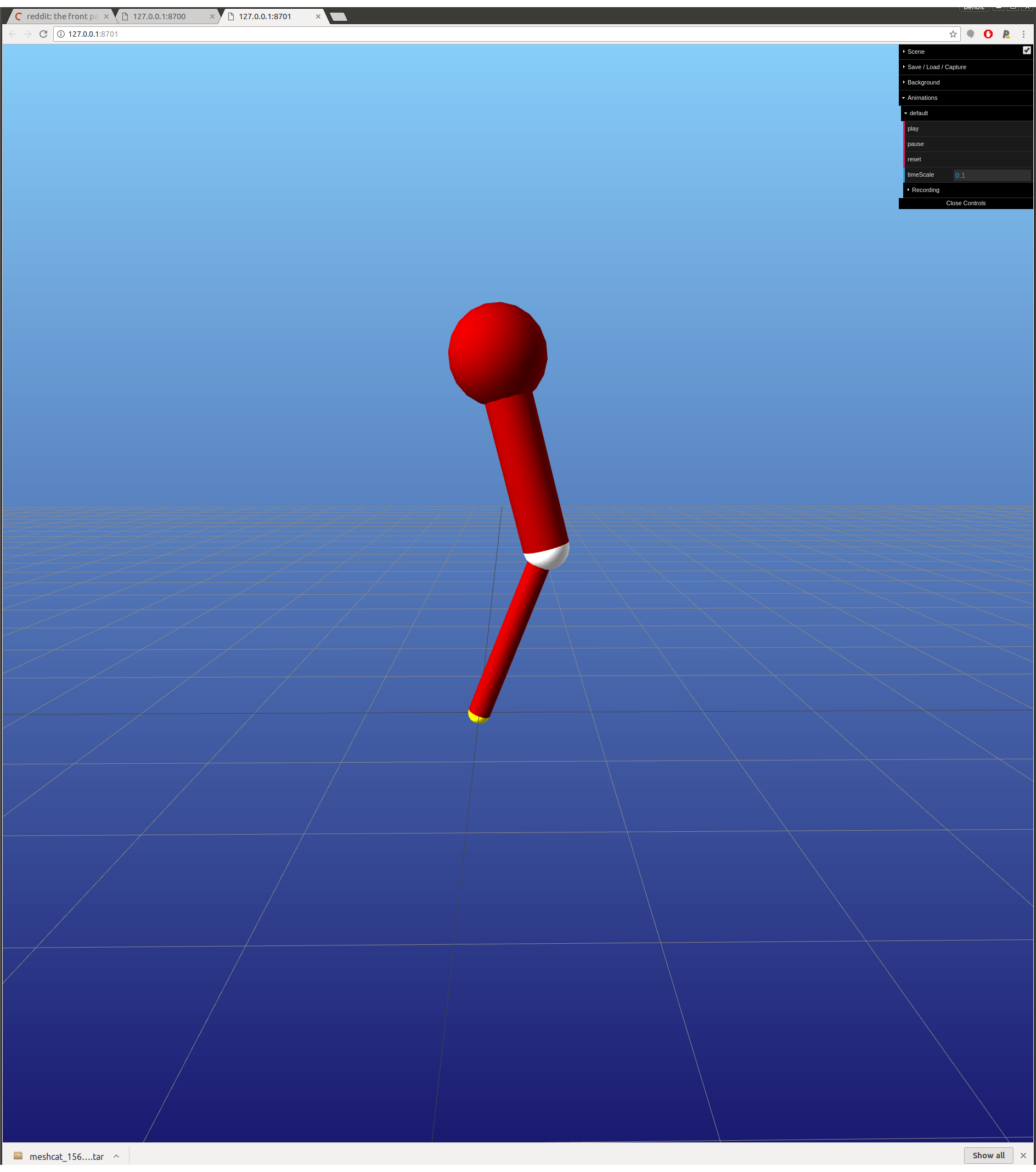}}
    \end{subfigure}
    \begin{subfigure}[t]{.19\textwidth}
        \frame{\includegraphics[trim={10cm 17cm 10cm 9cm},clip,width=\textwidth]{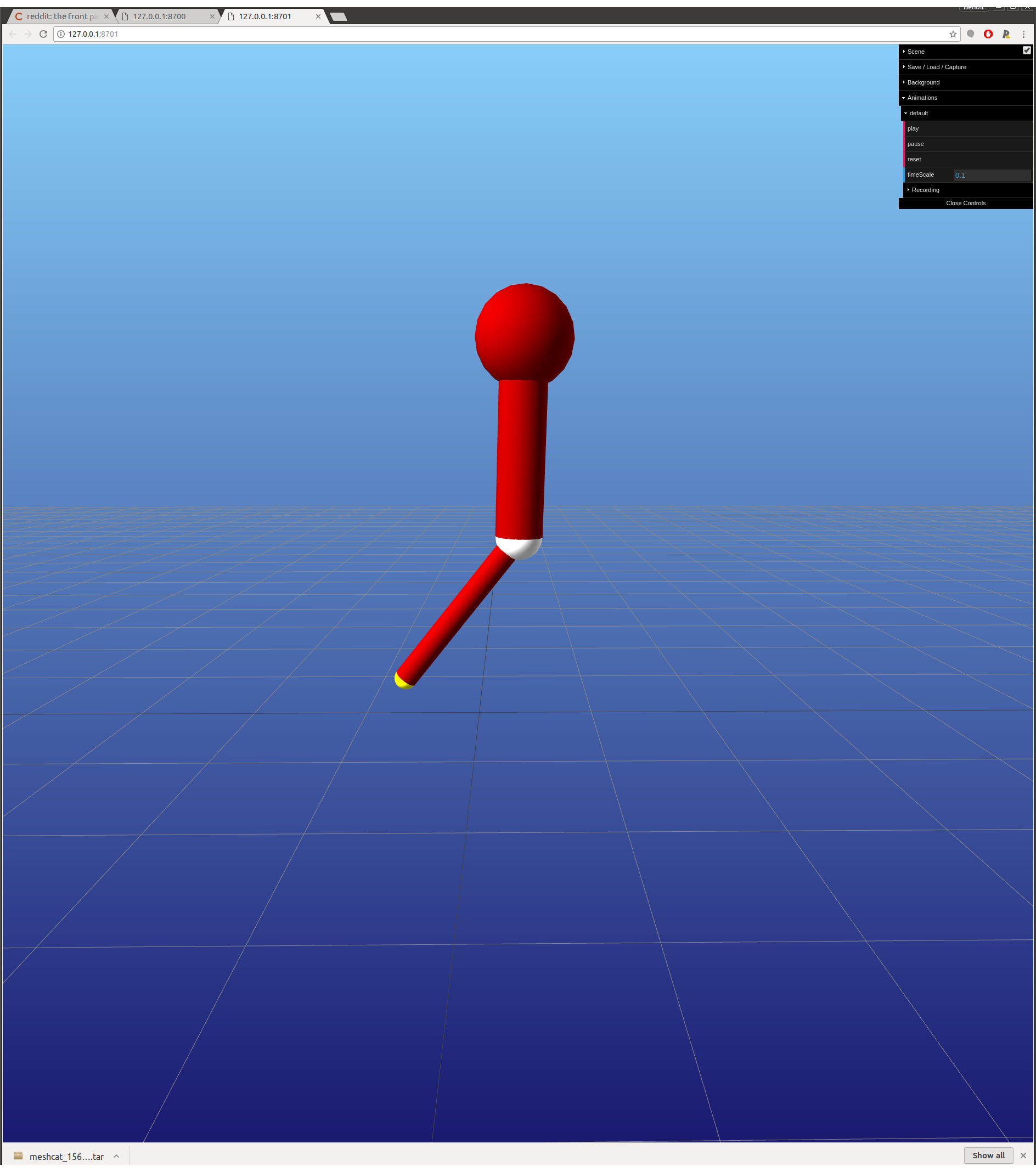}}
    \end{subfigure}
    \begin{subfigure}[t]{.19\textwidth}
        \frame{\includegraphics[trim={10cm 17cm 10cm 9cm},clip,width=\textwidth]{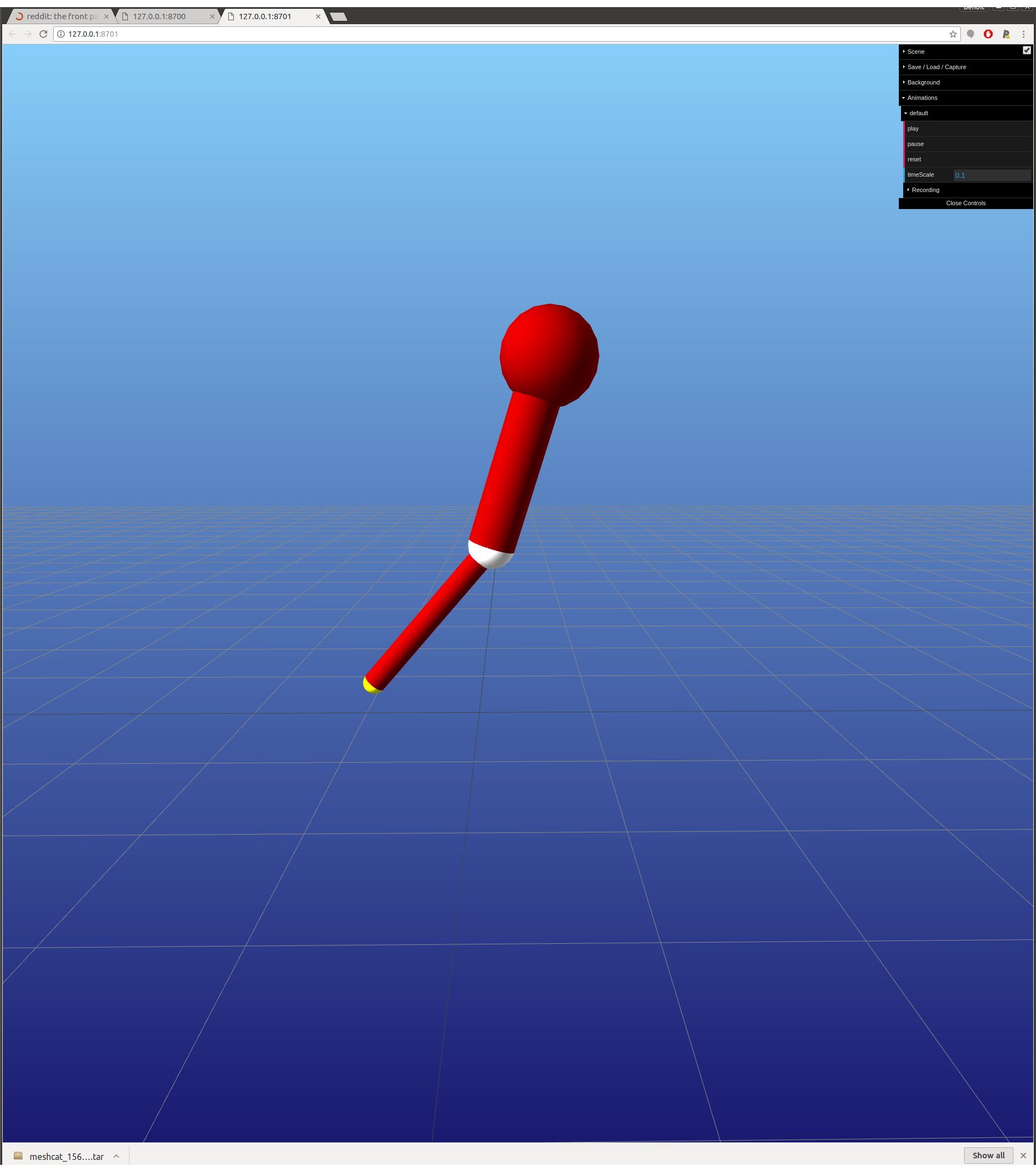}}
    \end{subfigure}
    \begin{subfigure}[t]{.19\textwidth}
        \frame{\includegraphics[trim={10cm 17cm 10cm 9cm},clip,width=\textwidth]{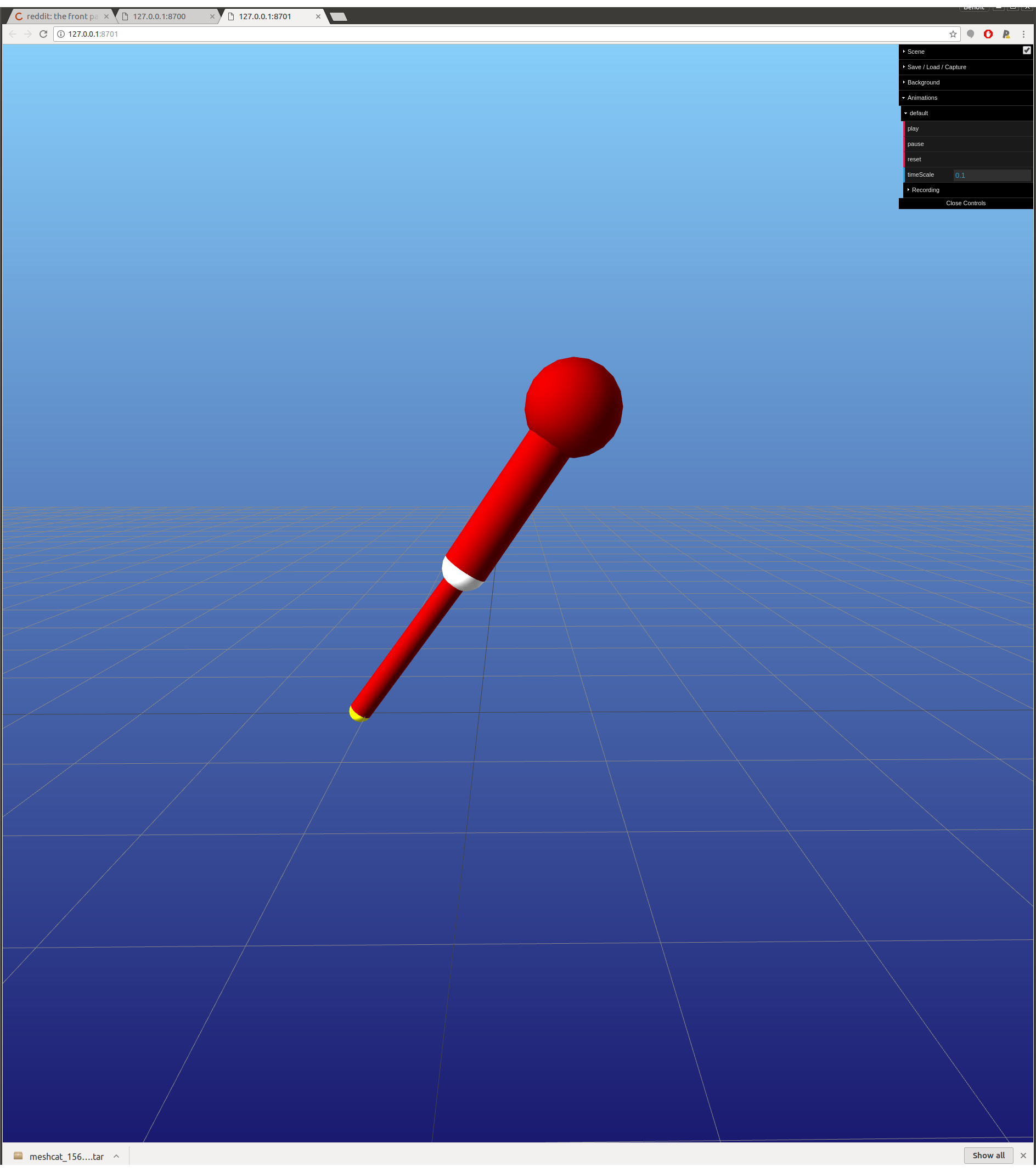}}
    \end{subfigure}
    \caption{One of the resulting trajectories of the additional benchmark task for computational time comparison. The goal is for a five degrees-of-freedom robot to jump to a specified height. Only the knee is actuated.}
    \label{fig:hoppingimg}
\end{figure*}

For thoroughness, we also benchmark a second different toy problem. This time, the problem consists of a hopping robot with a single actuator at the knee and a contact point at its foot. The system has 4 degrees of freedom and one input. The task is to jump to a target height from a given initial configuration, given actuator limits. One of the resulting trajectories is shown in figure \ref{fig:hoppingimg}. Like in the first benchmark, both methods were allowed to run until optimality (with a tolerance of $10^{-5}$). The resulting solve times in seconds are reported in table \ref{table:hoppingtable}. Once again, our semidirect method outperformed the indirect method.
\begin{table}[]
\centering
\begin{tabular}{|c|c|c|}
\hline
\textbf{\# Knot points} & \textbf{Indirect (s)} & \textbf{Semidirect (s) - ours} \\ \hline
10 & 0.152 & {\ul 0.126} \\ \hline
15 & 0.474 & {\ul 0.255} \\ \hline
20 & 0.678 & {\ul 0.374} \\ \hline
\end{tabular}
\vspace{0.25cm}
\caption{Additional benchmark task for computational time comparison with state-of-the-art alternative method. The task consists of getting a hopping robot to reach a target height as shown in \ref{fig:hoppingimg}.}
\label{table:hoppingtable}
\end{table}

\subsection{Parallelization Benchmark} \label{subsec:parallelres}

Here, we also report the additional reduction in mean run time that is possible to achieve with our semidirect method by evaluating the constraints (and therefore solving the lower problems) in parallel. The problems solved here are the same ones as described in section \ref{subsec:perfbench} (with their serial counterparts reported in \ref{tab:slideingbox}). A comparison is displayed in table \ref{table:parallel}. As expected, the computational gains of parallelization increase with problem size.
\begin{table}[]
\centering
\begin{tabular}{|c|c|c|}
\hline
\textbf{\# Knot points} & \textbf{\begin{tabular}[c]{@{}c@{}}Semidirect \\ serial (s)\end{tabular}} & \textbf{\begin{tabular}[c]{@{}c@{}}Semidirect \\ parallel (s)\end{tabular}} \\ \hline
10 & 0.054 & {\ul 0.031} \\ \hline
15 & 0.140 & {\ul 0.105} \\ \hline
20 & 0.229 & {\ul 0.193} \\ \hline
25 & 0.359 & {\ul 0.220} \\ \hline
30 & 0.524 & {\ul 0.293} \\ \hline
35 & 1.258 & {\ul 0.435} \\ \hline
\end{tabular}
\vspace{0.25cm}
\caption{Additional reduction in mean run-time (reported in seconds) of the sliding box benchmark from section \ref{subsec:perfbench} when constraints are evaluated (and therefore the lower problems solved) in parallel. As expected, computational gains increase with problem size.}
\label{table:parallel}
\end{table}

\subsection{Quadruped Gait Design} \label{subsec:littledog}

Next, we applied our approach to perform trajectory optimization for Boston Dynamics's ``little dog''. The system and the contact points are modeled as 3-dimensional, but the system is constrained such that its center of mass moves in a vertical plane. The system has 15 degrees of freedom (from the position of the center of mass, orientation in the plane, and leg joint angles) and each leg is modeled with a contact point on its tip.

\begin{figure*}[ht]
    \setlength{\fboxsep}{0pt}%
    \setlength{\fboxrule}{1pt}%
    \centering
    \begin{subfigure}[t]{.242\textwidth}
        \fbox{\includegraphics[trim={7cm 25cm 24cm 16cm},clip,width=\textwidth]{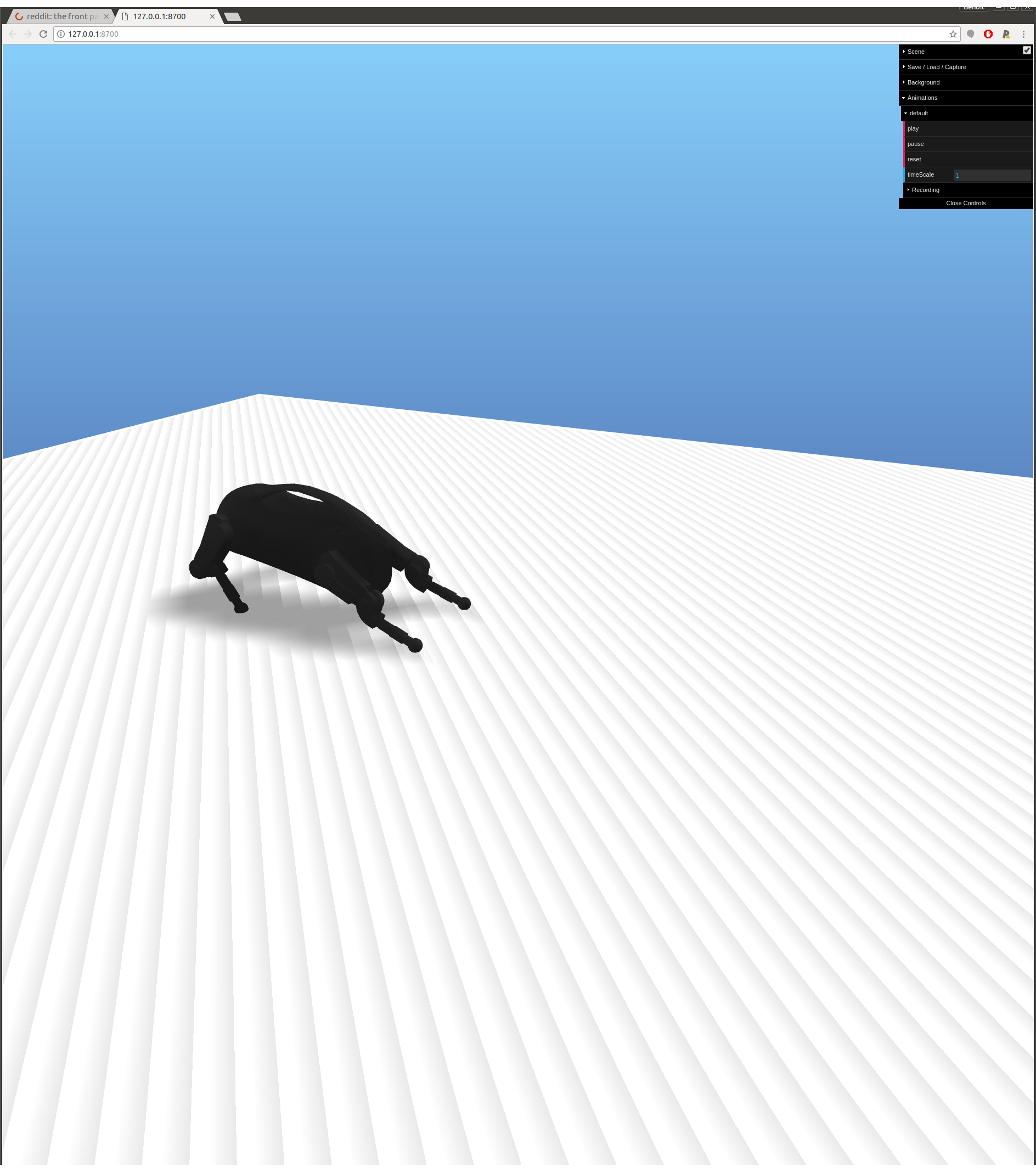}}
        \vspace{.5mm}
    \end{subfigure}
    \begin{subfigure}[t]{.242\textwidth}
        \fbox{\includegraphics[trim={7cm 25cm 24cm 16cm},clip,width=\textwidth]{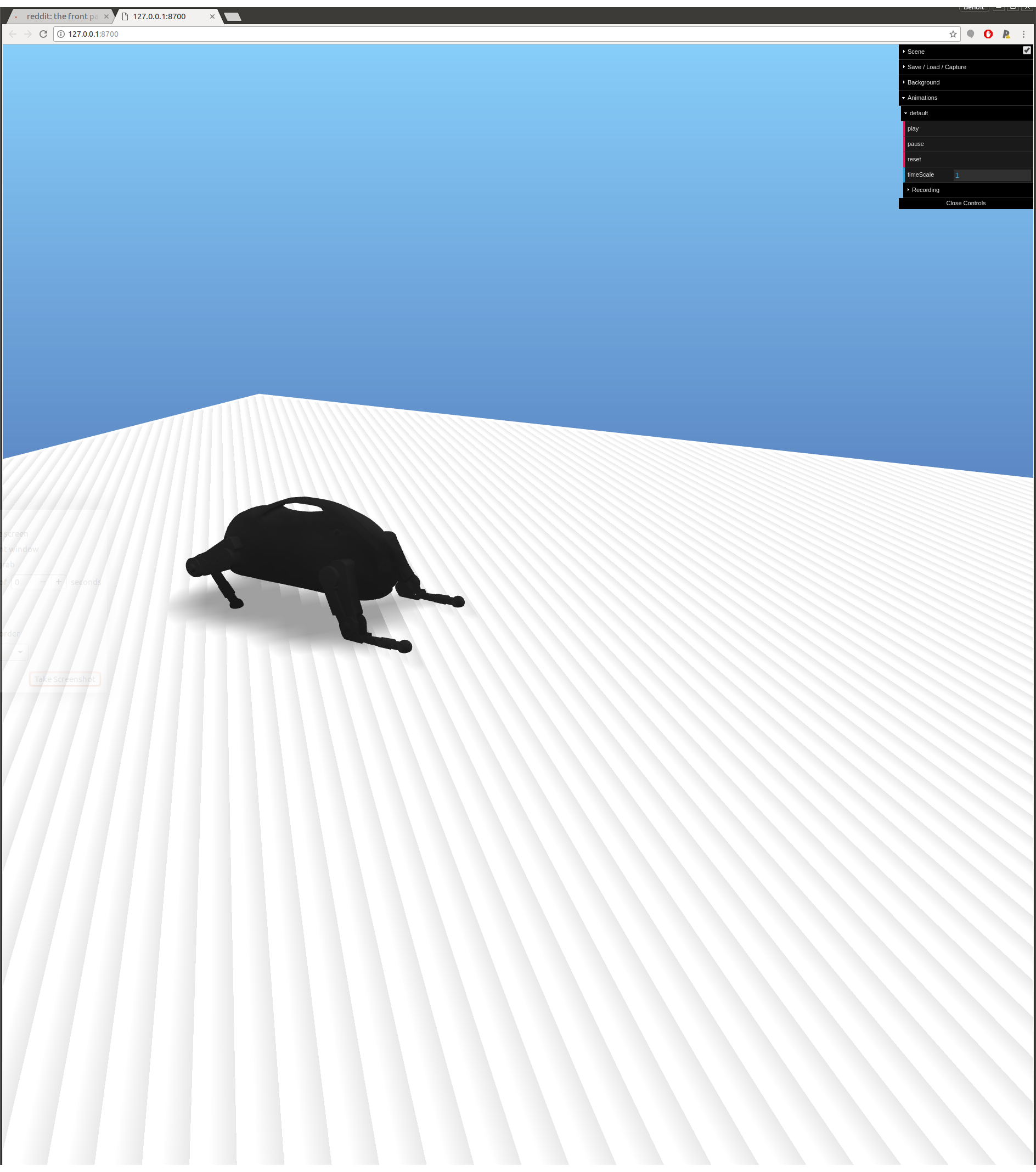}}
        \vspace{.5mm}
    \end{subfigure}
    \begin{subfigure}[t]{.242\textwidth}
        \fbox{\includegraphics[trim={7cm 25cm 24cm 16cm},clip,width=\textwidth]{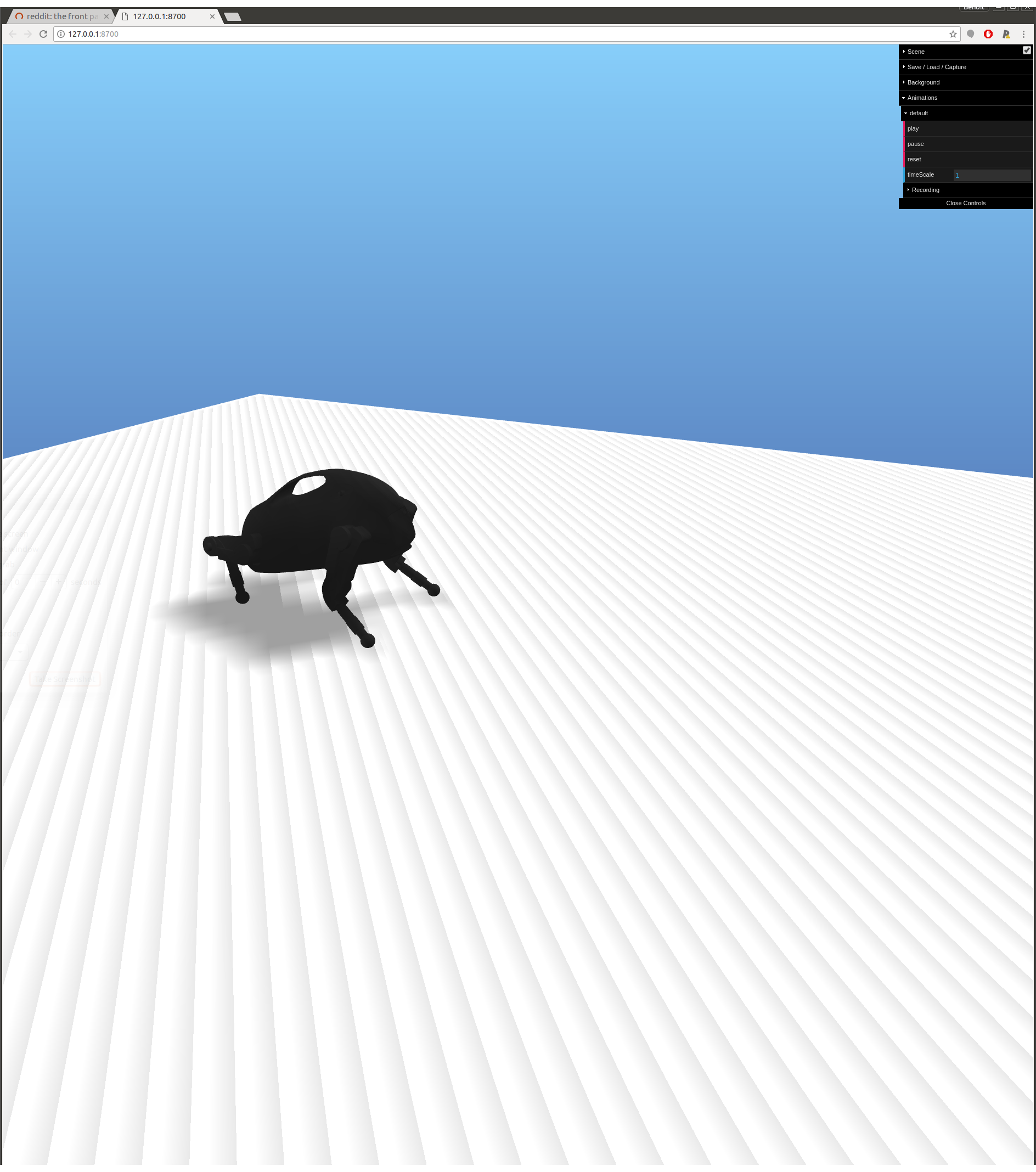}}
        \vspace{.5mm}
    \end{subfigure}
    \begin{subfigure}[t]{.242\textwidth}
        \fbox{\includegraphics[trim={7cm 25cm 24cm 16cm},clip,width=\textwidth]{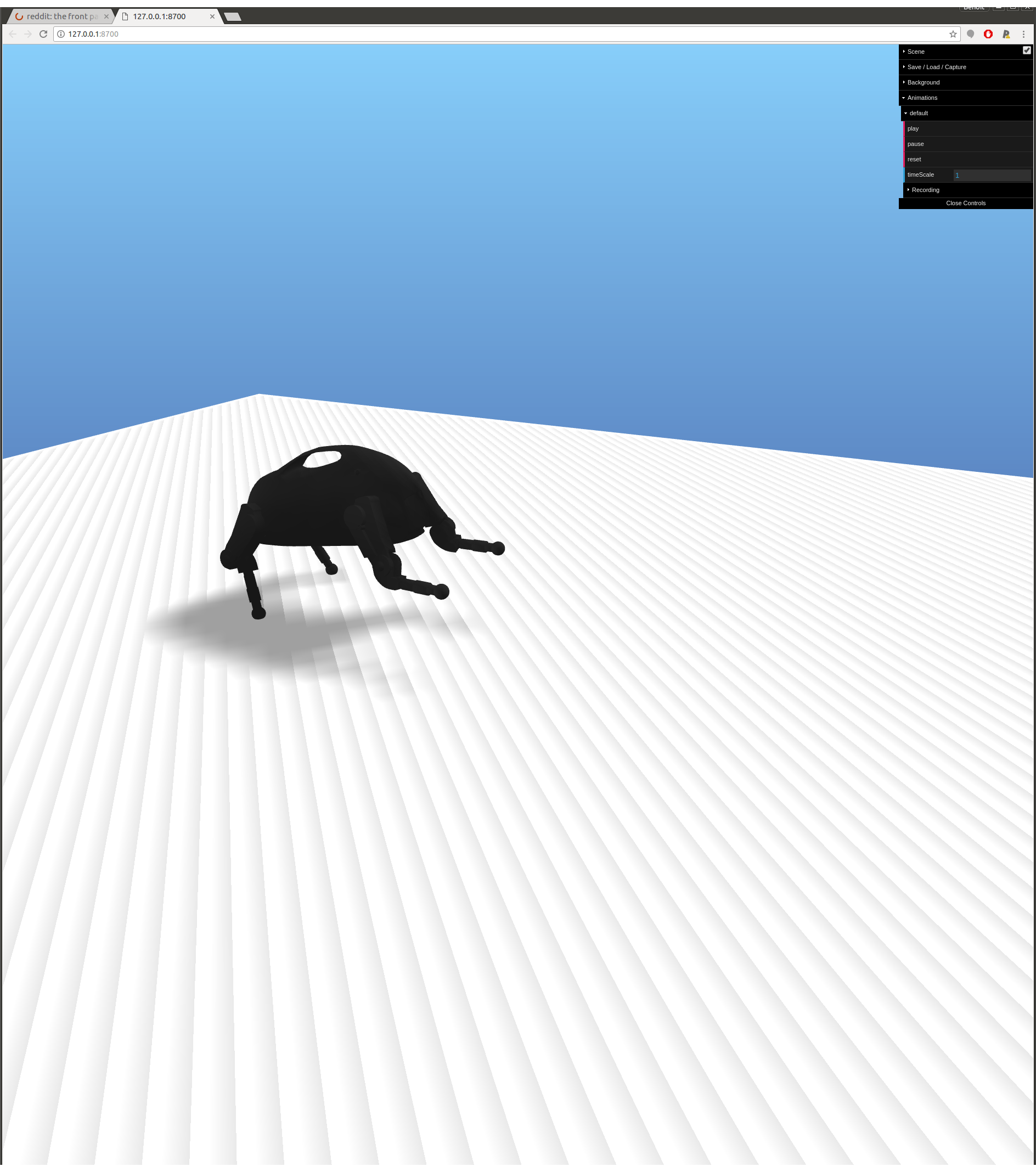}}
        \vspace{.5mm}
    \end{subfigure}
    \begin{subfigure}[t]{.242\textwidth}
        \fbox{\includegraphics[trim={7cm 25cm 24cm 16cm},clip,width=\textwidth]{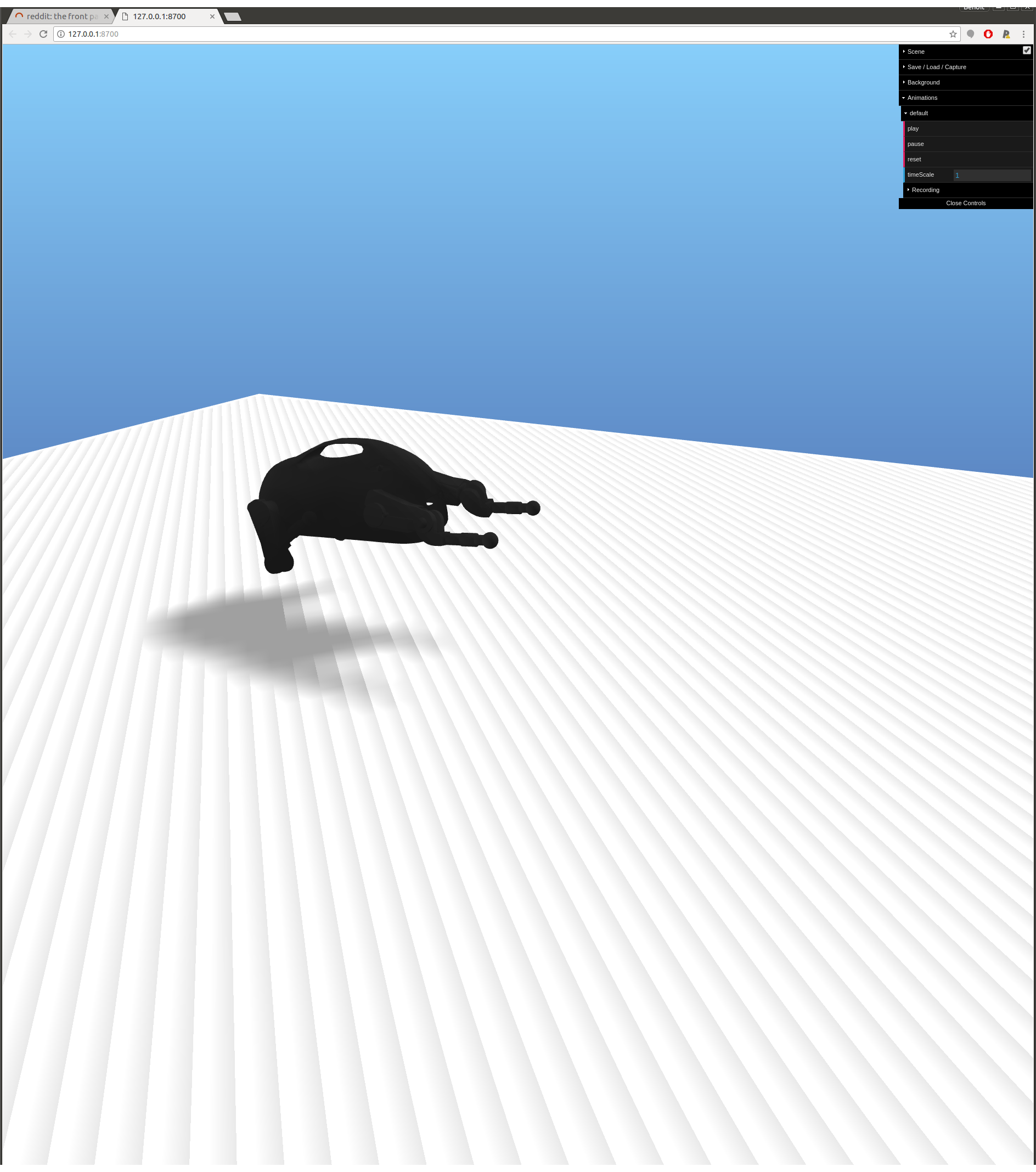}}
    \end{subfigure}
    \begin{subfigure}[t]{.242\textwidth}
        \fbox{\includegraphics[trim={7cm 25cm 24cm 16cm},clip,width=\textwidth]{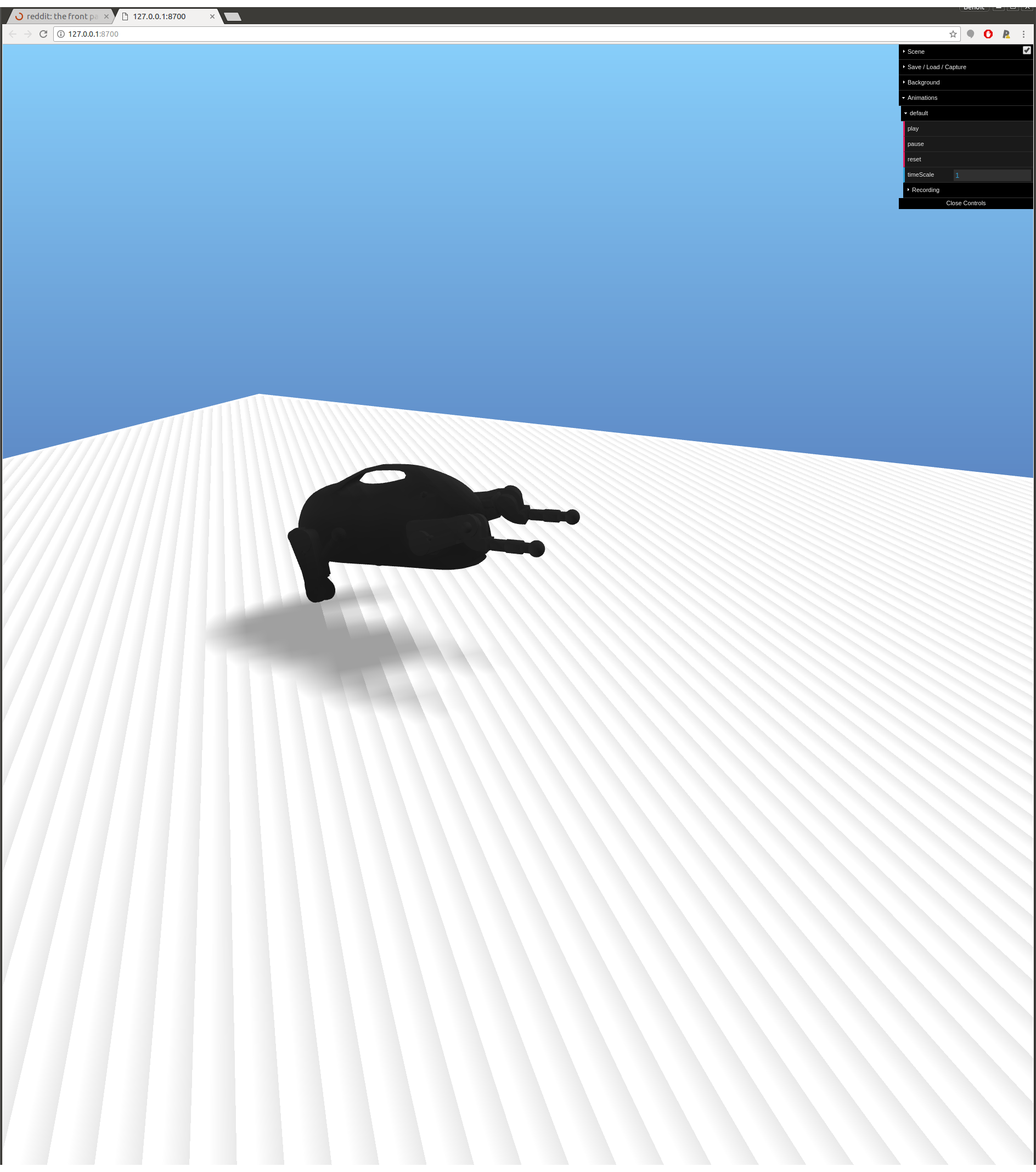}}
    \end{subfigure}
    \begin{subfigure}[t]{.242\textwidth}
        \fbox{\includegraphics[trim={7cm 25cm 24cm 16cm},clip,width=\textwidth]{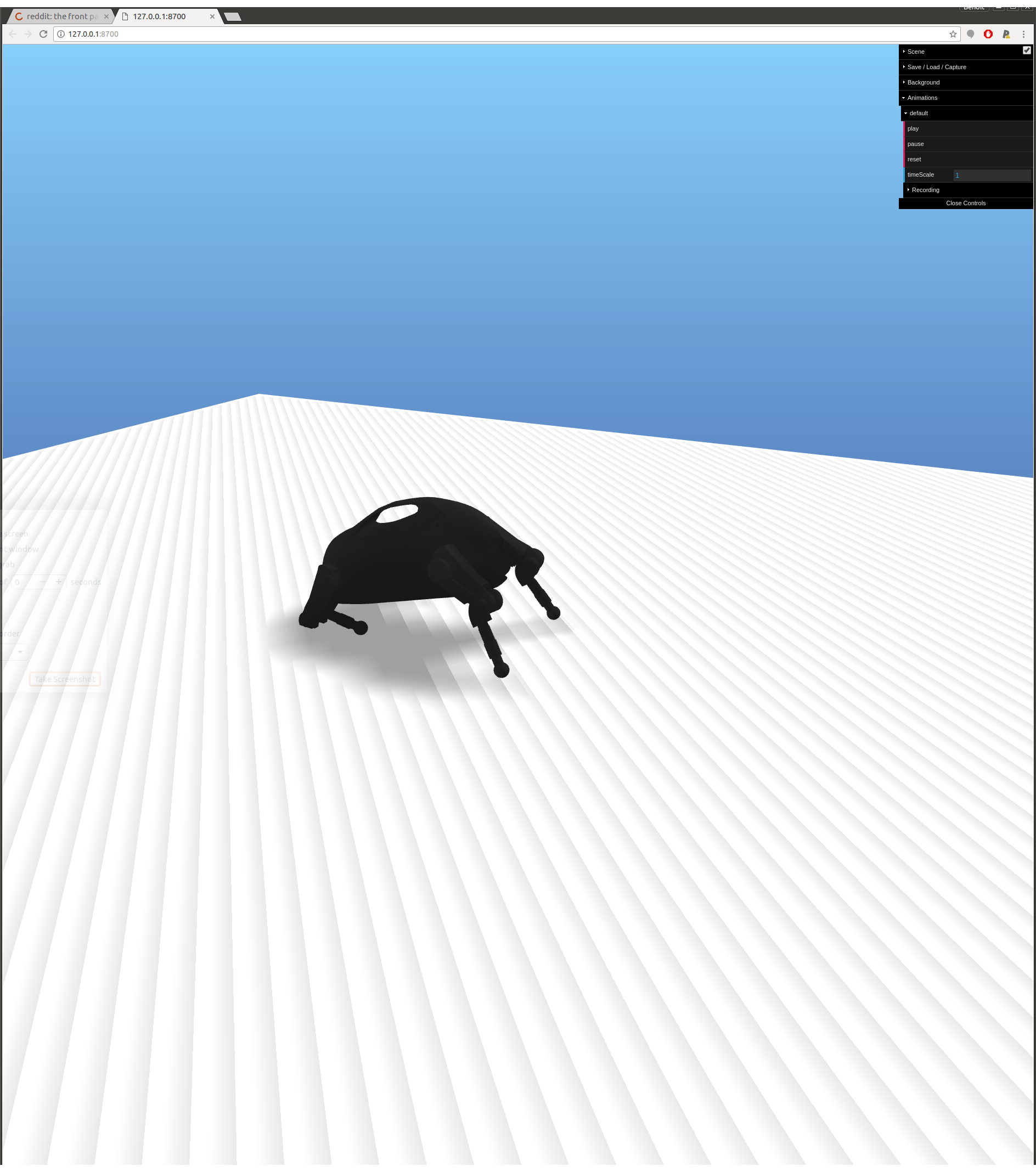}}
    \end{subfigure}
    \begin{subfigure}[t]{.242\textwidth}
        \fbox{\includegraphics[trim={7cm 25cm 24cm 16cm},clip,width=\textwidth]{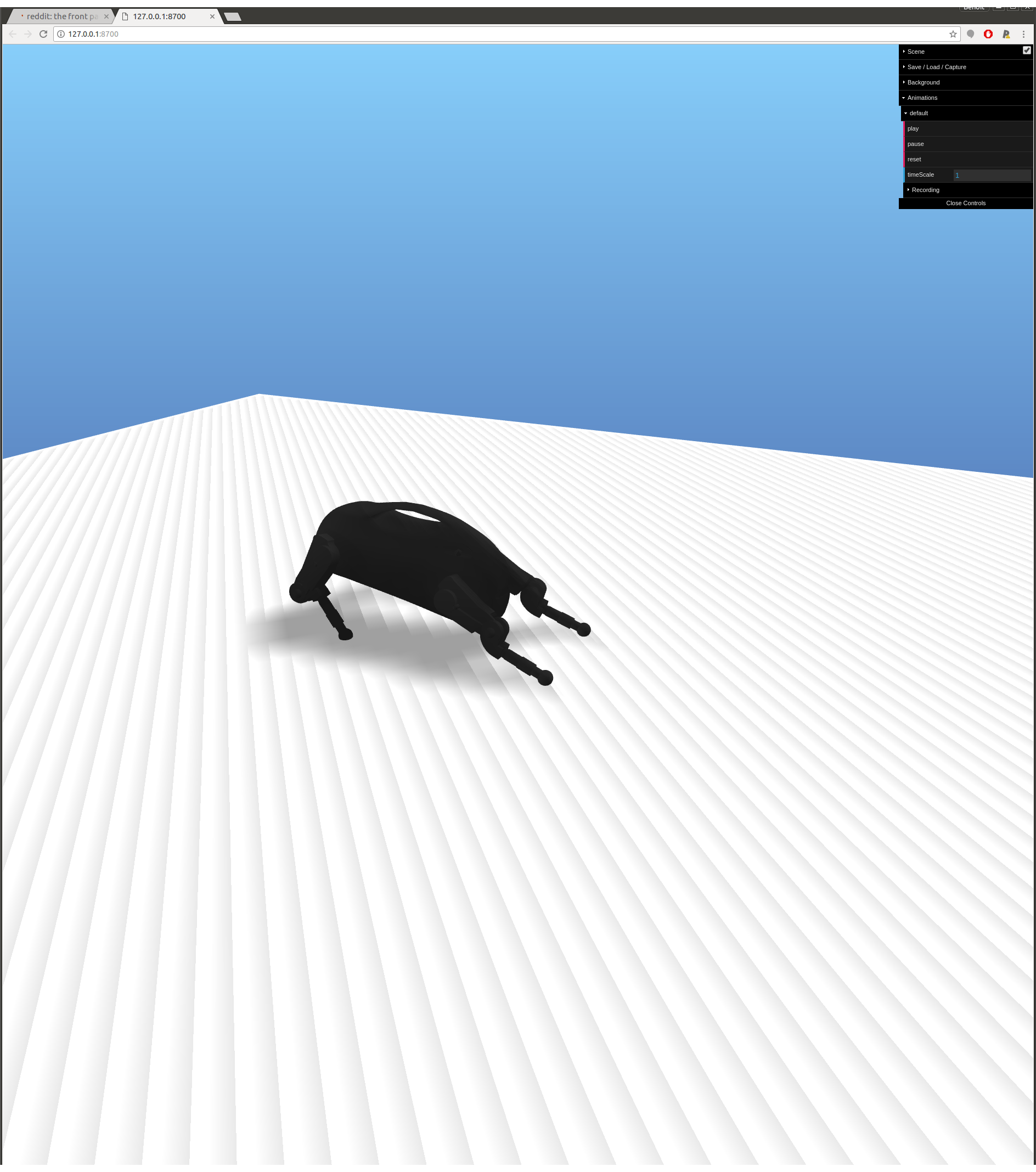}}
    \end{subfigure}
    \caption{The gait found by our semidirect method for a quadruped, little dog. The robot first lowers itself towards the ground and then quickly moves upwards while pushing backwards (leveraging implicit friction forces computed by the embedded optimization problem). This generates a forward ``flight`` phase followed by a landing phase where little dog brings itself to rest (once again leveraging friction with the ground) and returns to its initial configuration. This makes up one cycle of a gait that exploits complex interactions between the robot and its environment through friction and normal forces.}
    \label{fig:dogstep}
\end{figure*}

The problem consists of moving little dog forward by 20 centimeters, starting and ending with zero velocity, while  also respecting actuator upper and lower limits. In order to increase the trajectory's practical implications, we constraint it to describe a periodic gait by also enforcing that the final configuration (minus the forward displacement) matches the initial one. In Figure \ref{fig:dogstep} we show snapshots of the trajectory resulting from our proposed semidirect method, and a video of the resulting gait is also available at \url{https://www.youtube.com/playlist?list=PL8-2mtIlFIJpgmWImauC9rXxgN2wWpcIR}. 
% \begin{figure}
% \centering
% \begin{subfigure}{\textwidth}
%   \centering
%   \includegraphics[width=.95\linewidth]{step1.png}
% \end{subfigure}
% \begin{subfigure}{1\textwidth}
%   \centering
%   \includegraphics[width=.95\linewidth]{step2.png}
% \end{subfigure}

\begin{figure}
\centering
\begin{subfigure}{\textwidth}
  \centering
  \includegraphics[width=.9\linewidth]{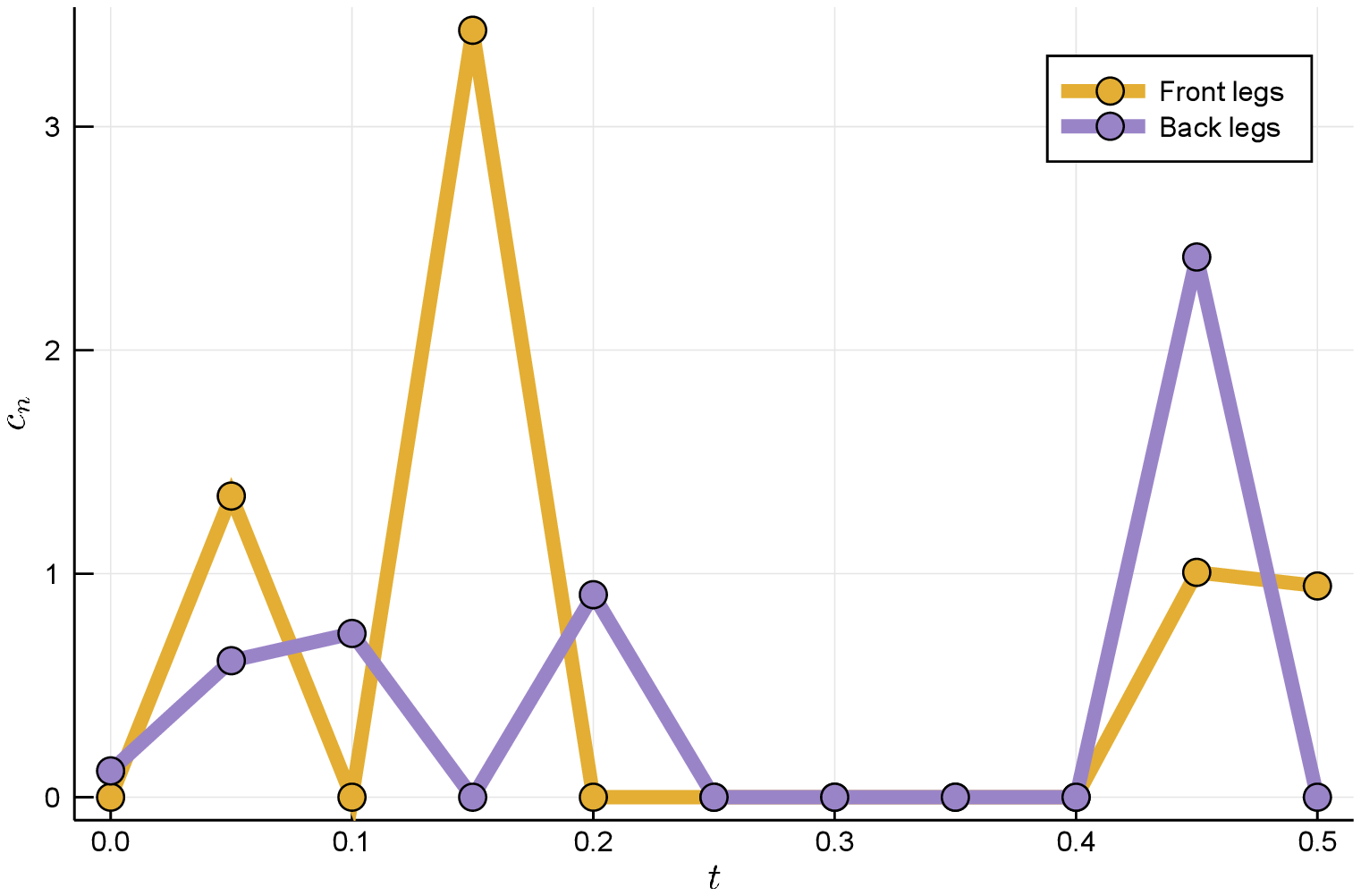}
  \vspace{1cm}
\end{subfigure}
\begin{subfigure}{1\textwidth}
  \centering
  \includegraphics[width=.9\linewidth]{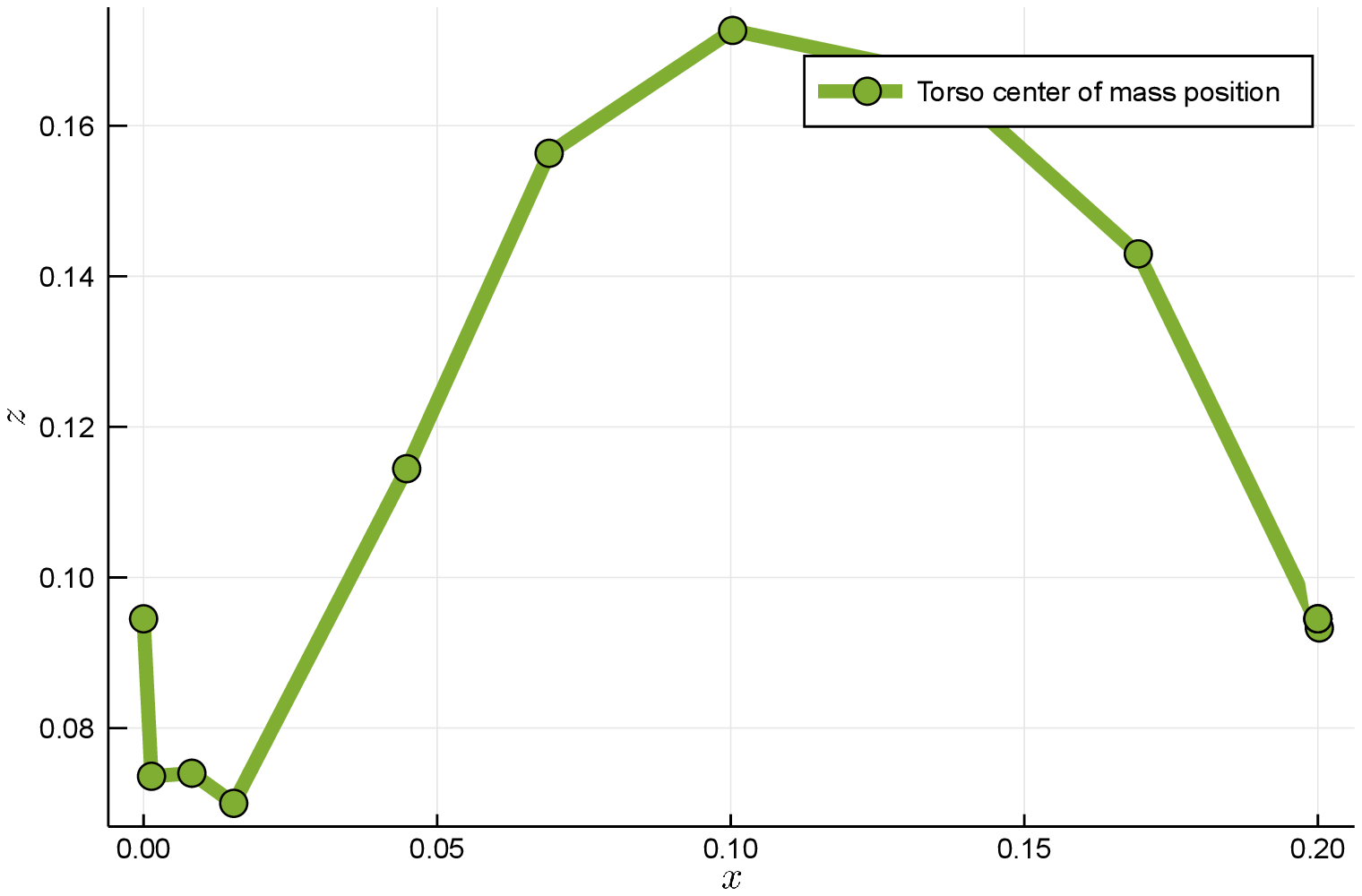}
\end{subfigure}
\caption{Normal contact forces at the legs (top) and position of the center of mass (bottom) of little dog during one cycle of the gait resulting from the trajectory optimization problem described in Section \ref{subsec:littledog}.}
\label{fig:dognormal}
\end{figure}

Figure \ref{fig:dognormal} shows the total normal force on the front and back legs, clearly demonstrating the non-trivial strategy that the algorithm found to design a gait. For this example, the trajectory optimization solve time (over several samples) is benchmarked at 1.8 seconds. After our best effort, we were unable to get the indirect method from \cite{PosaCantuEtAl2014} to converge to a good solution for this task and therefore cannot report its performance on it. However an alternative method such as \cite{DaiValenzuelaEtAl2014}, which \emph{neglects several dynamic constraints in order to make the optimization easier}, reports computation times similar to ours for the same quadruped.

\section{Method Shortcomings}
We note that even though our approach is capable of efficiently generating a vast array of complex trajectories that involve making and breaking contact, it still leaves many open questions. Notably our strategy (least squares) for choosing a subgradient when the gradient of the lower quadratic problem is not uniquely defined seems to work in practice, but does not rest on a strong theoretical foundation yet, and is perhaps far from optimal for this application. Moreover, in practice, we found that trajectories involving sticking contact seemed to pose a bigger challenge to the upper solver (SNOPT) that would sometimes struggle to improve the trajectory past a certain point, most likely due to the gradients of the lower problem. We believe these are great directions for future research in both theoretical and numerical aspects of our method.

\section{Conclusion}
In this work we introduced a bilevel optimization approach to robotic trajectory optimization for systems that can make and break contact with their environment. The approach falls under the category of planning ``through contact'', where the contact constraints are directly resolved within the formulated optimization problem. While similar approaches have been proposed in the past, the novelty of our proposed method is that it formulates the contact constraints in the trajectory optimization in a ``semidirect'' way. Specifically, the normal force contact is handled indirectly via complementarity constraints in the optimization problem and the friction force is handled directly as the solution to an embedded optimization problem. This allows us to avoid to use of additional complementarity constraints for the friction forces and to avoid linearizing the non-penetration constraints. To demonstrate empirically the advantages of our proposed approach we presented results from three problems: two benchmark problems involving sticking and sliding friction and a gait optimization for a 15-degrees-of-freedom quadruped.

\section*{Acknowledgments}
This work was supported in part by the Office of Naval Research (Grant N00014-17-1-2749). The authors of this work would also like to thank Hongkai Dai for his guidance at multiple stages of this work, as well as Shane Barratt for his useful pointers.

\bibliographystyle{unsrt}
\bibliography{main}

\end{document}